\documentclass{article} 

\usepackage[T1]{fontenc}
\usepackage{tgtermes}

\usepackage{amssymb}
\newcommand{\mathbold}[1]{\ensuremath{\boldsymbol{\mathbf{#1}}}}

\usepackage{times}

\usepackage{amsmath, amsthm, amsfonts}
\usepackage{mathtools}

\usepackage[
  paper  = letterpaper,
  left   = 1.25in,
  right  = 1.25in,
  top    = 1.0in,
  bottom = 1.0in,
  ]{geometry}
\usepackage[english]{babel}
\usepackage[parfill]{parskip}
\usepackage{afterpage}
\usepackage{enumitem}
\usepackage{framed}
\usepackage{xspace}

\usepackage{lineno}

\usepackage{ragged2e}

\newcounter{parcount}

\usepackage[usenames,dvipsnames]{xcolor}
\definecolor{shadecolor}{gray}{0.9}

\usepackage{graphicx}
\usepackage[labelfont=bf]{caption}
\usepackage[format=hang]{subcaption}

\usepackage{booktabs, array}

\usepackage[algoruled]{algorithm2e}
\usepackage{listings}
\usepackage{fancyvrb}
\fvset{fontsize=\small}

\usepackage{natbib}
\usepackage[colorlinks,linktoc=all]{hyperref}
\usepackage[all]{hypcap}
\hypersetup{citecolor=MidnightBlue}
\hypersetup{linkcolor=MidnightBlue}
\hypersetup{urlcolor=MidnightBlue}
\usepackage[nameinlink]{cleveref}
\creflabelformat{equation}{#2\textup{#1}#3}  

\usepackage
[acronym,smallcaps,nowarn,section,nonumberlist]{glossaries}
\glsdisablehyper{}

\definecolor{strings}{rgb}{.624,.251,.259}
\definecolor{keywords}{rgb}{.224,.451,.686}
\definecolor{comment}{rgb}{.322,.451,.322}

\lstdefinelanguage{python}{
  keywords=[3]{Normal, Bernoulli, Beta, Categorical, Dirichlet,
  Exponential, MultivariateNormalFull, RandomVariable,
  DirichletProcess, Empirical, PointMass, Gamma,
  MAP, Inference, KLqp, HMC, SGLD, KLpq,
  VariationalInference, MonteCarlo, ConjugateInference, GANInference,
  rnn_cell, dirichlet_process, cond, body, generative_network,
  discriminative_network,
  evaluate, ppc, copy, dot, get_session},
  morecomment=[l]{\#},
  morecomment=[s]{"""}{"""},
  morestring=[b]',
  morestring=[b]",
  alsoletter={<>=-+/*},
  sensitive=true
}

\renewcommand{\texttt}[1]{\lstinline[basicstyle=\fontsize{8pt}{8.25pt}\selectfont\ttfamily]{#1}}

\usepackage{iclr2017_conference}
\newacronym{PPL}{ppl}{probabilistic programming language}
\newacronym{GPU}{\textnormal{\uppercase{gpu}}}{graphics processing unit}

\newacronym{VAE}{vae}{variational auto-encoder}
\newacronym{RBM}{rbm}{restricted Boltzmann machine}
\newacronym{DLGM}{dlgm}{deep latent Gaussian model}
\newacronym{GAN}{gan}{generative adversarial network}
\newacronym{RNN}{rnn}{recurrent neural network}

\newacronym{VI}{vi}{variational inference}
\newacronym{MCMC}{mcmc}{Markov chain Monte Carlo}
\newacronym{MC}{mc}{Monte Carlo}
\newacronym{HMC}{hmc}{Hamiltonian Monte Carlo}
\newacronym{MAP}{map}{maximum a posteriori}
\newacronym{IWAE}{iwae}{importance-weighted auto-encoder}
\newacronym{HVM}{hvm}{hierarchical variational model}
\newacronym{KL}{kl}{Kullback-Leibler}
\newacronym{ELBO}{elbo}{\emph{evidence lower bound}}

\newcommand{\mba}{\mathbold{a}}
\newcommand{\mbb}{\mathbold{b}}

\newcommand{\mbh}{\mathbold{h}}

\newcommand{\mbp}{\mathbold{p}}

\newcommand{\mbx}{\mathbold{x}}
\newcommand{\mby}{\mathbold{y}}
\newcommand{\mbz}{\mathbold{z}}

\newcommand{\mbU}{\mathbold{U}}
\newcommand{\mbV}{\mathbold{V}}
\newcommand{\mbW}{\mathbold{W}}

\newcommand{\mbY}{\mathbold{Y}}

\newcommand{\mbepsilon}{\mathbold{\epsilon}}

\newcommand{\mblambda}{\mathbold{\lambda}}

\newcommand{\mbtheta}{\mathbold{\theta}}

\newcommand{\g}{\,|\,}
\renewcommand{\gg}{\,\|\,}

\usepackage{tikz}
\usetikzlibrary{bayesnet}

\pgfdeclarelayer{edgelayer}
\pgfdeclarelayer{nodelayer}
\pgfsetlayers{edgelayer,nodelayer,main}

\definecolor{hexcolor0xbfbfbf}{rgb}{0.749,0.749,0.749}

\tikzset{>=latex}
\tikzstyle{none}   = [inner sep=0pt]

\tikzstyle{line}  = [ - ]
\tikzstyle{arrow}  = [ ->, shorten <=1pt, shorten >=1pt ]
\tikzstyle{ardash} = [ dotted, ->, shorten <=1pt, shorten >=1pt ]

\tikzstyle{empty}=[circle,opacity=0.0,text opacity=1.0,inner sep=0pt,minimum
width=0pt,minimum height=0pt]
\tikzstyle{box}=[rectangle,fill=White,draw=Black]
\tikzstyle{filled}=[circle,fill=hexcolor0xbfbfbf,draw=Black]
\tikzstyle{hollow}=[circle,fill=White,draw=Black]
\tikzstyle{param}=[rectangle,fill=Black,draw=Black,inner sep=0pt,minimum width=4pt,minimum height=4pt]

\newcommand{\eat}[1]{}

\title{Deep Probabilistic Programming}

\author{%
Dustin Tran \\
Columbia University
\And
Matthew D. Hoffman \hspace{-0.7em} \\
Adobe Research
\And
Rif A. Saurous \hspace{8.65em} \\
Google Research
\AND
Eugene Brevdo \\
Google Brain
\And
Kevin Murphy \\
Google Research
\And
David M. Blei \\
Columbia University
}

\iclrfinalcopy

\begin{document}

\maketitle

\begin{abstract}
  We propose Edward, a Turing-complete \acrlong{PPL}. Edward defines
  two compositional representations---random variables and inference.
  By treating inference as a first class citizen, on a par with
  modeling, we show that probabilistic programming can be as
  flexible and computationally efficient as traditional deep learning.
  For flexibility,
  Edward makes it easy to fit the same model using a
  variety of composable inference methods, ranging from point
  estimation to variational inference to \acrshort{MCMC}.
  In addition, Edward can reuse the modeling representation as
  part of inference, facilitating the design of rich variational
  models and \acrlongpl{GAN}.
  For efficiency, Edward is integrated into
  TensorFlow, providing significant speedups over existing
  probabilistic systems. For example, we show on a benchmark logistic
  regression task that Edward is at least
  35x faster than Stan and 6x faster than PyMC3. Further, Edward
  incurs no runtime overhead: it is as fast as handwritten TensorFlow.
\end{abstract}

\vspace{-1ex}
\vspace{-1.0ex}
\section{Introduction}
\label{sec:introduction}
\vspace{-0.5ex}
The nature of deep neural networks is compositional. Users can connect
layers in creative ways, without having to worry about how to perform
testing (forward propagation) or inference (gradient-based
optimization, with back propagation and automatic differentiation).

In this paper, we design compositional representations for
probabilistic programming.  Probabilistic programming lets users
specify generative probabilistic models as programs and then
``compile'' those models down into inference procedures.
Probabilistic models are also compositional in nature, and much work
has enabled rich probabilistic programs via compositions of random
variables
\citep{goodman2012church,ghahramani2015probabilistic,lake2016building}.\\[0.75ex]
Less work, however, has considered an analogous compositionality for
inference. Rather, many existing \glsreset{PPL}\acrlongpl{PPL} treat
the inference engine as a black box, abstracted away from the model.
These cannot capture probabilistic inferences that
reuse the model's representation---a key idea in
recent advances in variational
inference~\citep{kingma2014autoencoding,rezende2015variational,tran2016variational},
\glsreset{GAN}\acrlongpl{GAN}~\citep{goodfellow2014generative},
and also in more classic inferences
\citep{dayan1995helmholtz,gutmann2010noise}.\\[0.75ex]
We propose Edward\footnote{%
  See \citet{tran2016edward}
  for details of the API. A companion webpage for this paper is available at
  \url{http://edwardlib.org/iclr2017}. It contains more complete
  examples with runnable code.}, a
Turing-complete \acrlong{PPL} which builds on two compositional
representations---one for random variables and one for inference.
By treating inference as a first class citizen, on a
par with modeling, we show that probabilistic programming can be as
flexible and computationally efficient as traditional deep learning.
For flexibility, we show how Edward makes it easy to fit
the same model using a variety of composable inference methods,
ranging from point estimation to variational inference to
\acrshort{MCMC}.
For efficiency, we
show how to integrate Edward into existing computational graph
frameworks such as TensorFlow \citep{abadi2016tensorflow}.  Frameworks
like TensorFlow provide computational benefits like distributed
training, parallelism, vectorization, and \glsunset{GPU}\gls{GPU}
support ``for free.''
For example, we show on a benchmark task that Edward's
\acrlong{HMC} is many times faster than existing software. Further, Edward
incurs no runtime overhead: it is as fast as handwritten TensorFlow.


\section{Related Work}
\label{sub:related}
\vspace{-0.5ex}

\Glspl{PPL} typically trade off the expressiveness of the language
with the computational efficiency of inference.  On one side, there
are languages which emphasize expressiveness
\citep{pfeffer2001ibal,milch2005blog,pfeffer2009figaro,goodman2012church},
representing a rich class beyond graphical models.  Each employs a
generic inference engine, but scales poorly with respect to model and
data size.  On the other side, there are languages which emphasize
efficiency
\citep{spiegelhalter1995bugs,murphy2001bayes,plummer2003jags,salvatier2015probabilistic,carpenter2016stan}.
The \gls{PPL} is restricted to a specific class of models, and
inference algorithms are optimized to be efficient for this class.
For example, Infer.NET enables fast message passing for graphical
models \citep{InferNET14}, and Augur enables data parallelism with
\glspl{GPU} for Gibbs sampling in Bayesian networks
\citep{tristan2014augur}.  Edward bridges this gap.  It is Turing
complete---it supports any computable probability distribution---and
it supports efficient algorithms, such as those that leverage model
structure and those that scale to massive data.

There has been some prior research on efficient algorithms in
Turing-complete languages.  Venture and Anglican design inference as
a collection of local inference problems, defined over program
fragments \citep{mansinghka2014venture,wood2014new}. This produces
fast program-specific inference code, which we build on.
Neither system supports inference methods such as programmable
posterior approximations, inference models, or data subsampling.
Concurrent with our work, WebPPL features amortized inference \citep{ritchie2016deep}.
Unlike Edward, WebPPL does not reuse the model's representation; rather, it annotates the
original program and leverages helper functions, which is a less
flexible strategy.  Finally, inference is designed as program
transformations in
\citet{kiselyov2009embedded,scibior2015practical,zinkov2016composing}.
This enables the flexibility of composing inference inside other
probabilistic programs. Edward builds on this idea to compose not only inference
within modeling but also modeling within inference (e.g.,
variational models).

\vspace{-0.5ex}
\section{Compositional Representations for Probabilistic Models}
\label{sec:modeling_language}
\vspace{-0.5ex}

We first develop compositional representations for probabilistic
models. We desire two criteria: (a) integration with computational
graphs, an efficient
framework where nodes represent operations on data and edges
represent data communicated between them
\citep{culler1986dataflow}; and (b)
invariance of the representation under the graph, that is, the
representation can be
reused during inference.

Edward defines random variables as the key compositional representation.
They are class objects with methods, for example, to compute the log
density and to sample. Further, each random variable $\mbx$ is
associated to a tensor (multi-dimensional array) $\mbx^*$, which
represents a single sample $\mbx^*\sim p(\mbx)$. This association
embeds the random variable onto a computational graph on tensors.

The design's simplicity makes it easy to develop probabilistic programs in a
computational graph framework. Importantly, all computation is
represented on the graph. This enables one to compose random
variables with complex deterministic structure such as deep neural
networks, a diverse set of math operations, and third party libraries
that build on the same framework. The design also enables compositions
of random variables to capture complex stochastic structure.

As an illustration, we use a Beta-Bernoulli model,
$p(\mbx, \theta) = \operatorname{Beta}(\theta\g 1, 1) \prod_{n=1}^{50}
\operatorname{Bernoulli}(x_n\g \theta)$, where $\theta$ is a latent
probability shared across the 50 data points $\mbx\in\{0,1\}^{50}$.
The random variable \texttt{x} is 50-dimensional, parameterized by the
random tensor $\theta^*$. Fetching the object \texttt{x} runs the
graph: it simulates from the generative process and outputs a binary
vector of $50$ elements.

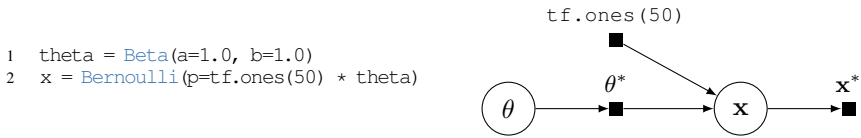
\begin{figure}[!htb]
\begin{subfigure}{0.3\columnwidth}
  \centering
\begin{lstlisting}[language=python]
theta = Beta(a=1.0, b=1.0)
x = Bernoulli(p=tf.ones(50) * theta)
\end{lstlisting}
\end{subfigure}%
\begin{subfigure}{0.65\columnwidth}
  \centering
  \begin{tikzpicture}[x=1.7cm,y=1.8cm,scale=0.9]

  \node[latent] (theta) {$\theta$};
  \factor[right=of theta, xshift=0.3cm] {thetastar} {$\theta^*$} {} {};

  \factor[above=of thetastar] {n} {\texttt{tf.ones(50)}} {} {};
  \node[latent, right=of thetastar, xshift=-0.5cm] (x) {$\mbx$};
  \factor[right=of x, xshift=0.3cm] {xstar} {$\mbx^*$} {} {};

  \edge{theta}{thetastar};
  \edge{thetastar}{x};
  \edge{n}{x};
  \edge{x}{xstar};

\end{tikzpicture}
\end{subfigure}
\caption{Beta-Bernoulli program \textbf{(left)} alongside its
computational graph \textbf{(right)}.
Fetching $\mbx$ from the graph generates a binary vector of $50$ elements.
}
\label{fig:beta_bernoulli}
\end{figure}

All computation is registered symbolically on random variables and not
over their execution. Symbolic representations
do not require reifying the full model, which leads to unreasonable
memory consumption for large models \citep{tristan2014augur}.
Moreover, it enables us to simplify both deterministic and stochastic
operations in the graph, before executing any code
\citep{scibior2015practical,zinkov2016composing}.

With computational graphs, it is also natural to build mutable states
within the probabilistic program. As a typical use of computational
graphs, such states can define model parameters; in TensorFlow, this
is given by a \texttt{tf.Variable}. Another use case is for building
discriminative models $p(\mby\g\mbx)$, where $\mbx$ are features that
are input as training or test data. The program can be written
independent of the data, using a mutable state
(\texttt{tf.placeholder}) for $\mbx$ in its graph. During training
and testing, we feed the placeholder the appropriate values.

In \Cref{appendix:model}, we provide examples of a Bayesian neural
network for classification (\ref{appendix:bnn}), latent Dirichlet
allocation (\ref{appendix:lda}), and Gaussian matrix factorization
(\ref{appendix:gaussian_mf}). We present others below.

\subsection{Example: Variational Auto-encoder}
\label{sub:vae}

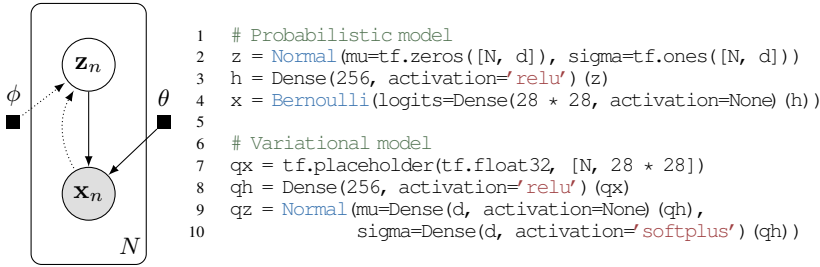
\begin{figure}[tb]
\begin{subfigure}{0.225\columnwidth}
  \centering
  \begin{tikzpicture}

  \node[latent] (z) {$\mbz_n$};
  \node[obs, below=of z] (x) {$\mbx_n$};

  \factor[empty, below=of z] {h} {} {} {};
  \factor[right=of h, xshift=0.5cm] {theta} {$\theta$} {} {};
  \factor[left=of h, xshift=-0.5cm] {phi} {$\phi$} {} {};

  \edge{z}{x};
  \draw[style=arrow, densely dotted, bend left] (x) to (z);
  \edge{theta}{x};
  \draw[style=arrow, densely dotted] (phi) to (z);

  \plate[inner sep=0.4cm, yshift=0.05cm,
    label={[xshift=-14pt,yshift=14pt]south east:$N$}] {plate1} {
    (z)(x)
  } {};

\end{tikzpicture}
  \label{sub:vae_math}
\end{subfigure}%
\begin{subfigure}{0.65\columnwidth}
  \centering
\begin{lstlisting}[language=python]
# Probabilistic model
z = Normal(mu=tf.zeros([N, d]), sigma=tf.ones([N, d]))
h = Dense(256, activation='relu')(z)
x = Bernoulli(logits=Dense(28 * 28, activation=None)(h))

# Variational model
qx = tf.placeholder(tf.float32, [N, 28 * 28])
qh = Dense(256, activation='relu')(qx)
qz = Normal(mu=Dense(d, activation=None)(qh),
            sigma=Dense(d, activation='softplus')(qh))
\end{lstlisting}
  \label{sub:vae_code}
\end{subfigure}
\caption{\Acrlong{VAE} for a data set of $28\times 28$ pixel images:
\textbf{(left)} graphical model, with dotted lines for the inference
model; \textbf{(right)} probabilistic program,
with 2-layer neural networks.
}
\label{fig:vae}
\end{figure}

\Cref{fig:vae} implements a \gls{VAE}
\citep{kingma2014autoencoding,rezende2014stochastic} in Edward. It
comprises a probabilistic model over data and a variational model
designed to approximate the former's posterior.
Here we use random variables to construct both
the probabilistic model and the variational model; they are fit
during inference (more details in \Cref{sec:inference}).

There are $N$ data points $x_n\in\{0,1\}^{28\cdot 28}$ each with
$d$ latent variables, $z_n\in\mathbb{R}^d$. The program uses Keras
\citep{chollet2015keras} to define neural networks. The
probabilistic model is parameterized by a 2-layer neural network, with
256 hidden units (and ReLU activation), and generates
$28\times 28$ pixel images. The variational model is parameterized by
a 2-layer inference network, with 256 hidden units and
outputs parameters of a normal posterior approximation.

The probabilistic program is concise. Core elements of the
\gls{VAE}---such as its distributional assumptions and neural net
architectures---are all extensible. With model compositionality, we
can embed it into more complicated models
\citep{gregor2015draw,rezende2016one} and for other learning tasks
\citep{kingma2014semi}. With inference compositionality (which we
discuss in \Cref{sec:inference}), we can embed it into more complicated algorithms, such
as with expressive variational approximations
\citep{rezende2015variational,tran2016variational,kingma2016improving}
and alternative objectives
\citep{ranganath2016operator,li2016variational,dieng2016chi}.

\subsection{\hspace{-0.225em}Example: Bayesian Recurrent Neural Network with Variable Length}

\begin{figure}[tb]
\begin{subfigure}{0.35\columnwidth}
  \centering
  \begin{tikzpicture}

  \node[empty]              (dot)      {} ;
  \node[obs, right=of dot, xshift=-0.75cm]        (xt) {$\mbx_t$} ;
  \node[latent, left=of xt, xshift=0.75cm, yshift=-0.75cm]  (bh)  {$\mbb_h$} ;
  \node[latent, above=of bh, yshift=-0.9cm]      (Wx) {$\mbW_x$} ;
  \node[latent, above=of Wx, yshift=-0.9cm]      (Wh)      {$\mbW_h$} ;
  \node[latent, below=of bh, yshift=0.9cm]      (Wy)      {$\mbW_y$} ;
  \node[latent, below=of Wy, yshift=1.0cm]      (by)      {$\mbb_y$} ;

  \factor[below=0.75cm of xt] {ht} {} {} {};
  \node[right=0.03cm of ht, yshift=-0.3cm] (htteyt) {$\mbh_t$};
  \node[empty, left=of ht]      (htminus)    {$\cdots$} ;
  \node[empty, right=of ht]      (htplus)    {$\cdots$} ;

  \node[obs, below=0.75cm of ht]      (yt)    {$\mby_t$} ;
  \node[empty, left=of yt]      (ytminus)    {} ;
  \node[empty, right=of yt]      (ytplus)    {} ;

  \edge{Wh}{ht};
  \edge{Wx}{ht};
  \edge{bh}{ht};
  \edge{xt}{ht};
  \edge{Wy}{yt};
  \edge{by}{yt};
  \edge{ht}{yt};
  \edge{htminus}{ht};
  \edge{ht}{htplus};

\end{tikzpicture}
\end{subfigure}%
\begin{subfigure}{0.6\columnwidth}
\begin{lstlisting}[language=python]
def rnn_cell(hprev, xt):
  return tf.tanh(tf.dot(hprev, Wh) + tf.dot(xt, Wx) + bh)

Wh = Normal(mu=tf.zeros([H, H]), sigma=tf.ones([H, H]))
Wx = Normal(mu=tf.zeros([D, H]), sigma=tf.ones([D, H]))
Wy = Normal(mu=tf.zeros([H, 1]), sigma=tf.ones([H, 1]))
bh = Normal(mu=tf.zeros(H), sigma=tf.ones(H))
by = Normal(mu=tf.zeros(1), sigma=tf.ones(1))

x = tf.placeholder(tf.float32, [None, D])
h = tf.scan(rnn_cell, x, initializer=tf.zeros(H))
y = Normal(mu=tf.matmul(h, Wy) + by, sigma=1.0)
\end{lstlisting}
\end{subfigure}
\caption{Bayesian \glsunset{RNN}\gls{RNN}: \textbf{(left)} graphical model;
  \textbf{(right)} probabilistic program. The program has an unspecified number
  of time steps; it uses a symbolic for loop (\texttt{tf.scan}). }
\label{fig:bayesian_rnn}
\end{figure}
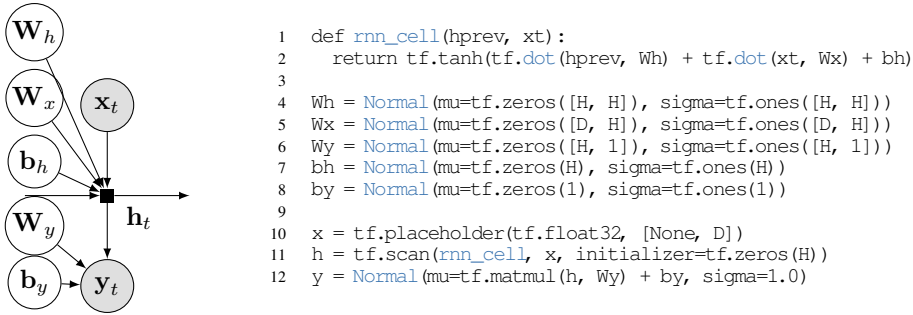

Random variables can also be composed with control flow operations.
As an example, \Cref{fig:bayesian_rnn} implements a Bayesian \glsreset{RNN}\gls{RNN} with
variable length.
The data is a sequence of inputs $\{\mbx_1,\ldots,\mbx_T\}$ and
outputs $\{y_1,\ldots,y_T\}$ of length $T$ with
$\mbx_t\in\mathbb{R}^{D}$ and $y_t\in\mathbb{R}$ per time step.
For $t=1,\ldots,T$,
a \gls{RNN} applies the update
\begin{equation*}
  \mbh_t = \operatorname{tanh}(\mbW_h \mbh_{t-1} + \mbW_x \mbx_t + \mbb_h),
\end{equation*}
where the previous hidden state is
$\mbh_{t-1}\in\mathbb{R}^H$.
We feed each hidden state into the output's likelihood,
$y_t \sim \operatorname{Normal}(\mbW_y \mbh_t + \mbb_y, 1)$, and
we place a standard normal prior over all parameters
$\{\mbW_h\in\mathbb{R}^{H\times H}, \mbW_x\in\mathbb{R}^{D\times H},
\mbW_y\in\mathbb{R}^{H\times 1},
\mbb_h\in\mathbb{R}^H,\mbb_y\in\mathbb{R}\}$. Our implementation is
dynamic: it differs from a \gls{RNN} with fixed length, which
pads and unrolls the computation.

\subsection{Stochastic Control Flow and Model Parallelism}

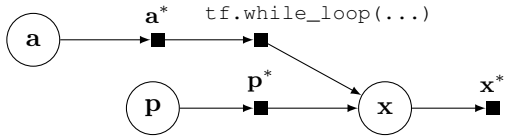
\begin{figure}[!htb]
  \centering
  \begin{tikzpicture}[x=1.7cm,y=1.8cm,scale=0.9]

  \node[latent] (p) {$\mbp$};
  \factor[right=of p, xshift=0.3cm] {pstar} {$\mbp^*$} {} {};

  \factor[above=of pstar] {n} {} {} {};
  \factor[empty, right=of n, yshift=0.1cm] {nn} {\texttt{tf.while_loop(...)}} {} {};
  \factor[left=of n, xshift=-0.5cm] {astar} {$\mba^*$} {} {};
  \node[latent, left=of astar, xshift=0.5cm] (a) {$\mba$};
  \node[latent, right=of pstar, xshift=-0.5cm] (x) {$\mbx$};
  \factor[right=of x, xshift=0.3cm] {xstar} {$\mbx^*$} {} {};

  \edge{p}{pstar};
  \edge{pstar}{x};
  \edge{n}{x};
  \edge{x}{xstar};
  \edge{a}{astar};
  \edge{astar}{n};

\end{tikzpicture}
\caption{Computational graph for a probabilistic program with stochastic control flow.
}
\label{fig:dynamic}
\end{figure}

Random variables can also be placed in the control flow itself,
enabling probabilistic programs with stochastic control flow.
Stochastic control flow defines dynamic conditional dependencies,
known in the literature as contingent or existential dependencies
\citep{mansinghka2014venture,wu2016swift}. See \Cref{fig:dynamic},
where $\mbx$ may or may not depend on $\mba$ for a given execution.
In \Cref{appendix:dirichlet_process}, we use stochastic control flow
to implement a Dirichlet process mixture model.
Tensors with stochastic shape are also possible: for
example, \texttt{tf.zeros(Poisson(lam=5.0))} defines a vector of zeros
with length given by a Poisson draw with rate $5.0$.

Stochastic control flow produces difficulties for algorithms that use
the graph structure because the relationship of conditional
dependencies changes across execution traces. The computational
graph, however, provides an elegant way of teasing out static
conditional dependence structure ($\mbp$) from dynamic dependence
structure ($\mba)$. We can perform model parallelism (parallel
computation across components of the model) over the static structure
with \glspl{GPU} and batch training. We can use more generic
computations to handle the dynamic structure.

\section{Compositional Representations for Inference}
\label{sec:inference}

We described random variables as a representation for building
rich probabilistic programs over computational graphs.  We now
describe a compositional representation for inference.
We desire two criteria: (a) support for many classes of
inference, where the form of the inferred posterior depends on the
algorithm; and (b) invariance of inference under the computational
graph, that is, the posterior can be further composed as part of
another model.

\begin{figure}[!h]
\begin{subfigure}{0.35\columnwidth}
  \centering
  \begin{tikzpicture}

  \node[latent]               (beta)      {$\beta$} ;
  \node[left=0.4cm of beta]               (betaph)      {} ;
  \node[latent, below=1.0cm of betaph] (z)    {$\mbz_n$} ;
  \node[obs, right=1.0cm of z]      (x)        {$\mbx_n$} ;

  \edge{beta}{x};
  \edge{z}{x};

  \plate[inner sep=0.3cm,
    label={[xshift=-15pt,yshift=15pt]south east:$N$}] {plate1} {
    (x)(z)
  } {};

\end{tikzpicture}
\end{subfigure}%
\begin{subfigure}{0.6\columnwidth}
  \centering
\begin{lstlisting}[language=python]
N = 10000  # number of data points
D = 2  # data dimension
K = 5  # number of clusters

beta = Normal(mu=tf.zeros([K, D]), sigma=tf.ones([K, D]))
z = Categorical(logits=tf.zeros([N, K]))
x = Normal(mu=tf.gather(beta, z), sigma=tf.ones([N, D]))
\end{lstlisting}
\end{subfigure}
\caption{Hierarchical model: \textbf{(left)} graphical model; \textbf{(right)}
  probabilistic program. It is a mixture of Gaussians over
  $D$-dimensional data $\{x_n\}\in\mathbb{R}^{N\times D}$. There are
  $K$ latent cluster means $\beta\in\mathbb{R}^{K\times D}$.}
\label{fig:hierarchical_model_example}
\end{figure}
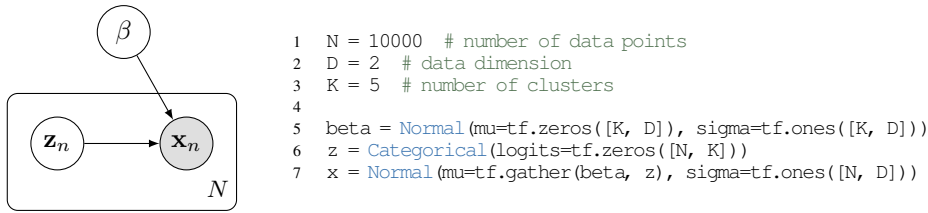

To explain our approach, we will use a simple hierarchical model as a
running example. \Cref{fig:hierarchical_model_example} displays a
joint distribution $p(\mbx, \mbz, \beta)$ of data $\mbx$, local
variables $\mbz$, and global variables $\beta$. The ideas here extend
to more expressive programs.

\subsection{Inference as Stochastic Graph Optimization}
\label{sub:inference}

The goal of inference is to
calculate the posterior distribution
$p(\mathbf{z}, \beta\mid \mathbf{x}_{\text{train}}; \mbtheta)$
given data $\mbx_{\text{train}}$,
where
$\mbtheta$ are any model parameters that we will compute point estimates
for.\footnote{%
For example, we could replace \texttt{x}'s \texttt{sigma}
argument with \texttt{tf.exp(tf.Variable(0.0))*tf.ones([N, D])}. This
defines a model parameter initialized at 0 and positive-constrained.}
We formalize this as the following optimization problem:
\begin{align}
  \label{eq:inference-optimization}
\min_{\mblambda,\mbtheta}
\mathcal{L}(
p(\mathbf{z}, \beta\mid \mathbf{x}_{\text{train}}; \mbtheta),~
q(\mathbf{z}, \beta; \mblambda)
),
\end{align}
where $q(\mathbf{z}, \beta; \mblambda)$ is an approximation to the
posterior $p(\mathbf{z}, \beta\g \mbx_{\text{train}};\mbtheta)$, and
$\mathcal{L}$ is a loss function with respect to $p$ and $q$.

The choice of approximation $q$, loss $\mathcal{L}$, and rules to update
parameters $\{\mbtheta,\mblambda\}$ are specified by an inference algorithm.
(Note $q$ can be nonparametric, such as a point or a collection of
samples.)

In Edward, we write this problem as follows:

\begin{lstlisting}[language=python]
inference = ed.Inference({beta: qbeta, z: qz}, data={x: x_train})
\end{lstlisting}

\texttt{Inference} is an abstract class which takes two inputs.  The
first is a collection of latent random variables \texttt{beta} and
\texttt{z}, associated to their ``posterior variables'' \texttt{qbeta} and
\texttt{qz} respectively. The second is a collection of observed random variables
\texttt{x}, which is associated to their realizations \texttt{x_train}.

The idea is that \texttt{Inference} defines and
solves the optimization in \Cref{eq:inference-optimization}. It
adjusts parameters of the distribution of \texttt{qbeta}
and \texttt{qz} (and any model parameters) to be close to the
posterior.

Class methods are available to finely control the inference. Calling
\texttt{inference.initialize()} builds a computational graph to update
$\{\mbtheta,\mblambda\}$. Calling \texttt{inference.update()} runs
this computation once to update $\{\mbtheta,\mblambda\}$; we call the
method in a loop until convergence. Importantly, no efficiency is lost
in Edward's language: the computational graph is the same as if it
were handwritten for a
specific model. This means the runtime is the same; also see our
experiments in \Cref{sub:gpu}.

A key concept in Edward is that there is no distinct ``model''
or ``inference'' block. A model is simply a collection of random
variables, and inference is a way of modifying parameters in that
collection subject to another. This reductionism offers
significant flexibility. For example, we can
infer only parts of a model (e.g.,
layer-wise training \citep{hinton2006fast}),
infer parts used in multiple models
(e.g., multi-task learning), or
plug in a posterior into a new model
(e.g., Bayesian updating).

\subsection{Classes of Inference}

The design of \texttt{Inference} is very general.  We describe
subclasses to represent many algorithms below: variational inference,
Monte Carlo, and \acrlongpl{GAN}.

\begin{figure}[!h]
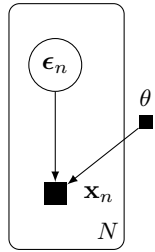

\begin{subfigure}{0.5\columnwidth}
  \centering
\begin{lstlisting}[language=Python]
qbeta = Normal(
  mu=tf.Variable(tf.zeros([K, D])),
  sigma=tf.exp(tf.Variable(tf.zeros([K, D]))))
qz = Categorical(
  logits=tf.Variable(tf.zeros([N, K])))

inference = ed.VariationalInference(
  {beta: qbeta, z: qz}, data={x: x_train})
\end{lstlisting}
\end{subfigure}%
\begin{subfigure}{0.5\columnwidth}
  \centering
\begin{lstlisting}[language=Python]
T = 10000  # number of samples
qbeta = Empirical(
  params=tf.Variable(tf.zeros([T, K, D])))
qz = Empirical(
  params=tf.Variable(tf.zeros([T, N])))

inference = ed.MonteCarlo(
  {beta: qbeta, z: qz}, data={x: x_train})
\end{lstlisting}
\end{subfigure}
\caption{\textbf{(left)} Variational inference. \textbf{(right)} Monte Carlo.}
\label{fig:inference}
\end{figure}

Variational inference posits a family of approximating distributions
and finds the closest member in the family to the posterior
\citep{jordan1999introduction}.  In Edward, we build the variational
family in the graph; see \Cref{fig:inference} (left). For our running
example, the
family has mutable variables as parameters
$\mblambda=\{\pi,\mu,\sigma\}$, where
$q(\beta;\mu,\sigma) = \operatorname{Normal}(\beta; \mu,\sigma)$ and
$q(\mbz;\pi) = \operatorname{Categorical}(\mbz;\pi)$.

Specific variational algorithms inherit from the
\texttt{VariationalInference} class.  Each defines its own methods,
such as a loss function and gradient.
For example, we represent
\gls{MAP}
estimation with an approximating family (\texttt{qbeta} and
\texttt{qz}) of \texttt{PointMass} random variables, i.e., with all
probability mass concentrated at a point.
\texttt{MAP} inherits from \texttt{VariationalInference} and defines
the negative log joint density as the loss function; it uses existing optimizers inside
TensorFlow.
In \Cref{sub:recent}, we experiment with multiple gradient estimators
for black box variational inference \citep{ranganath:2014}. Each
estimator implements the same loss (an objective proportional to
the divergence $\operatorname{KL}(q\gg p)$) and a different update rule
(stochastic gradient).

Monte Carlo approximates the posterior using samples
\citep{robert1999monte}. Monte Carlo is an inference where the
approximating family is an empirical distribution,
$q(\beta; \{\beta^{(t)}\}) = \frac{1}{T}\sum_{t=1}^T \delta(\beta,
\beta^{(t)})$ and
$q(\mbz; \{\mbz^{(t)}\}) = \frac{1}{T}\sum_{t=1}^T \delta(\mbz,
\mbz^{(t)})$. The parameters are
$\mblambda=\{\beta^{(t)},\mbz^{(t)}\}$.  See \Cref{fig:inference}
(right).
Monte Carlo algorithms proceed by updating one sample
$\beta^{(t)},\mbz^{(t)}$ at a time in the empirical approximation.
Specific \glsunset{MC}\gls{MC} samplers determine the update rules:
they can use gradients such as in Hamiltonian Monte Carlo
\citep{neal2011mcmc} and graph
structure such as in sequential Monte Carlo \citep{doucet2001introduction}.

\begin{figure}[tb]
\begin{subfigure}{0.4\columnwidth}
  \centering
  \begin{tikzpicture}

  \node[latent] (eps0) {$\mbepsilon_n$};
  \factor[minimum size=0.3cm, below=1.2cm of eps0] {x} {} {} {};

  \factor[empty, below=of eps0] {h} {} {} {};
  \factor[right=of h, xshift=0.7cm] {theta} {$\theta$} {} {};

  \node[right=of x, xshift=-0.9cm] (xlabel) {$\mbx_n$};

  \edge{eps0}{x};
  \edge{theta}{x};

  \plate[inner sep=0.4cm, yshift=0.05cm,xshift=0.15cm,
    label={[xshift=-14pt,yshift=14pt]south east:$N$}] {plate1} {
    (eps0)(x)
  } {};

\end{tikzpicture}
\end{subfigure}%
\begin{subfigure}{0.6\columnwidth}
  \centering
\begin{lstlisting}[language=python]
def generative_network(eps):
  h = Dense(256, activation='relu')(eps)
  return Dense(28 * 28, activation=None)(h)

def discriminative_network(x):
  h = Dense(28 * 28, activation='relu')(x)
  return Dense(h, activation=None)(1)

# Probabilistic model
eps = Normal(mu=tf.zeros([N, d]), sigma=tf.ones([N, d]))
x = generative_network(eps)

inference = ed.GANInference(data={x: x_train},
    discriminator=discriminative_network)
\end{lstlisting}
\end{subfigure}
\caption{\Acrlongpl{GAN}:
\textbf{(left)} graphical model; \textbf{(right)} probabilistic program.
The model (generator) uses a parameterized function (discriminator)
for training.
}
\label{fig:gan}
\end{figure}

Edward also supports non-Bayesian methods such as \glspl{GAN}
\citep{goodfellow2014generative}.
See \Cref{fig:gan}.
The model posits
random noise \texttt{eps} over $N$ data points, each with $d$
dimensions; this random noise feeds into a
\texttt{generative_network} function, a neural network that outputs
real-valued data \texttt{x}.
In addition, there is a \texttt{discriminative_network}
which takes data as input and outputs the probability that
the data is real (in logit parameterization). We build
\texttt{GANInference}; running it optimizes parameters inside the two
neural network functions. This approach extends to many advances in
\glspl{GAN} (e.g., \citet{denton2015deep,li2015generative}).

Finally, one can design algorithms that would otherwise require tedious
algebraic manipulation. With symbolic algebra on nodes of the
computational graph, we can uncover conjugacy relationships between
random variables. Users can then integrate out variables to
automatically derive classical Gibbs \citep{gelfand1990sampling},
mean-field updates \citep{bishop2006pattern}, and exact inference.
These algorithms are being currently developed in Edward.

\subsection{Composing Inferences}

Core to Edward's design is that inference can be written as a collection
of separate inference programs. Below we demonstrate variational EM,
with an (approximate) E-step over local variables and an M-step over
global variables. We instantiate two algorithms, each of which
conditions on inferences from the other, and
we alternate with one update of each \citep{neal1993new},
\begin{lstlisting}[language=Python]
qbeta = PointMass(params=tf.Variable(tf.zeros([K, D])))
qz = Categorical(logits=tf.Variable(tf.zeros([N, K])))

inference_e = ed.VariationalInference({z: qz}, data={x: x_train, beta: qbeta})
inference_m = ed.MAP({beta: qbeta}, data={x: x_train, z: qz})
...
for _ in range(10000):
  inference_e.update()
  inference_m.update()
\end{lstlisting}
This extends to many other cases such as
exact EM for exponential families,
contrastive divergence \citep{hinton2002training},
pseudo-marginal methods \citep{andrieu2009pseudo},
and Gibbs sampling within variational inference
\citep{wang2012truncation,Hoffman:2015}.
We can also write message passing algorithms, which solve a collection
of local inference problems \citep{koller2009probabilistic}.  For
example, classical message passing uses exact local inference and
expectation propagation locally minimizes the Kullback-Leibler
divergence, $\text{KL}(p\gg q)$
\citep{minka2001expectation}.

\subsection{Data Subsampling}
\label{sub:batch_training}

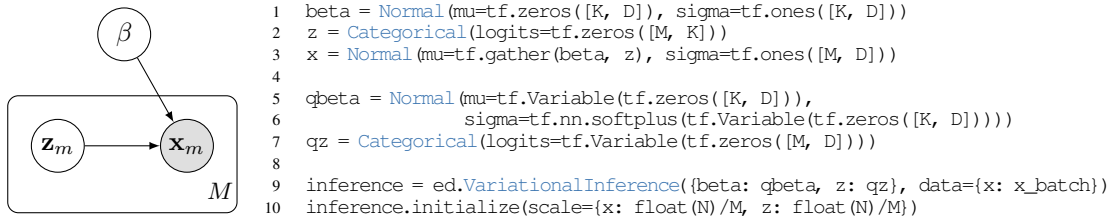
\begin{figure}[!htb]
\begin{subfigure}{0.3\columnwidth}
  \centering
  \begin{tikzpicture}

  \node[latent]               (beta)      {$\beta$} ;
  \node[left=0.4cm of beta]               (betaph)      {} ;
  \node[latent, below=1.0cm of betaph] (z)    {$\mbz_m$} ;
  \node[obs, right=1.0cm of z]      (x)        {$\mbx_m$} ;

  \edge{beta}{x};
  \edge{z}{x};

  \plate[inner sep=0.3cm,
    label={[xshift=-15pt,yshift=15pt]south east:$M$}] {plate1} {
    (x)(z)
  } {};

\end{tikzpicture}
  \label{sub:hierachical_model_math}
\end{subfigure}%
\begin{subfigure}{0.6\columnwidth}
  \centering
\begin{lstlisting}[language=python]
beta = Normal(mu=tf.zeros([K, D]), sigma=tf.ones([K, D]))
z = Categorical(logits=tf.zeros([M, K]))
x = Normal(mu=tf.gather(beta, z), sigma=tf.ones([M, D]))

qbeta = Normal(mu=tf.Variable(tf.zeros([K, D])),
               sigma=tf.nn.softplus(tf.Variable(tf.zeros([K, D]))))
qz = Categorical(logits=tf.Variable(tf.zeros([M, D])))

inference = ed.VariationalInference({beta: qbeta, z: qz}, data={x: x_batch})
inference.initialize(scale={x: float(N)/M, z: float(N)/M})
\end{lstlisting}
  \label{sub:hierarchical_model_code}
\end{subfigure}
\caption{Data subsampling with a hierarchical model. We define a
subgraph of the full model, forming a plate of size $M$
rather than $N$. We then scale all local random variables by $N/M$.}
\label{fig:hierachical_model_batch}
\end{figure}

Stochastic optimization \citep{bottou2010large} scales inference to
massive data and is key to
algorithms such as stochastic gradient Langevin dynamics
\citep{welling2011bayesian} and stochastic variational inference
\citep{hoffman2013stochastic}.  The idea is to cheaply estimate the
model's log joint density in an unbiased way.  At each step, one
subsamples a data set $\{x_m\}$ of size $M$ and then scales densities
with respect to local variables,
\begin{align*}
  \log p(\mbx, \mbz, \beta)
  & = \log p(\beta) +
  \sum_{n=1}^N\Big[\log p(x_n \g z_n, \beta) + \log p(z_n \g \beta)\Big]
  \\
  & \approx \log p(\beta) +
  \frac{N}{M}\sum_{m=1}^M\Big[\log p(x_m \g z_m, \beta) + \log p(z_m \g \beta)\Big].
\end{align*}
To support stochastic optimization, we represent only a subgraph of
the full model. This prevents reifying the full model, which can lead
to unreasonable memory consumption \citep{tristan2014augur}.  During
initialization, we pass in a dictionary to properly scale the
arguments. See \Cref{fig:hierachical_model_batch}.

Conceptually, the scale argument represents scaling for each random
variable's plate, as if we had seen that random variable $N / M$ as
many times.  As an example, \Cref{appendix:svi} shows how to implement
stochastic variational inference in Edward.
The approach extends naturally to streaming data
\citep{doucet2000on,broderick2013streaming,mcinerney2015population},
dynamic batch sizes, and data structures in which working on a
subgraph does not immediately apply
\citep{binder1997space,johnson2014stochastic,foti2014stochastic}.

\section{Experiments}
\label{sec:experiments}

In this section, we illustrate two main benefits of Edward:
flexibility and efficiency. For the former, we
show how it is easy to compare different inference algorithms on the
same model. For the latter, we show how it is easy to get significant
speedups by exploiting computational graphs.

\subsection{Recent Methods in Variational Inference}
\label{sub:recent}

\begin{table}[tb]
\centering
\begin{tabular}{lcc}
\toprule
Inference method & Negative log-likelihood
\\
\midrule
\gls{VAE} \citep{kingma2014autoencoding} & $\le$ 88.2 \\
\gls{VAE} without analytic KL & $\le$ 89.4 \\
\gls{VAE} with analytic entropy & $\le$ 88.1 \\
\gls{VAE} with score function gradient & $\le$ 87.9 \\
Normalizing flows \citep{rezende2015variational} & $\le$ 85.8 \\
Hierarchical variational model \citep{ranganath2016hierarchical} & $\le$ 85.4 \\
Importance-weighted auto-encoders ($K=50$) \citep{burda2016importance}
& $\le$ 86.3 \\
\acrshort{HVM} with \acrshort{IWAE} objective ($K=5$)
& $\le$ 85.2 \\
R\'{e}nyi divergence ($\alpha=-1$) \citep{li2016variational}
& $\le$ 140.5 \\
\bottomrule
\end{tabular}
\caption{Inference methods for a probabilistic decoder on binarized
MNIST. The Edward \gls{PPL} is a convenient research platform, making
it easy to both develop and experiment with many algorithms.}
\label{table:mnist}
\end{table}

We demonstrate Edward's flexibility for experimenting with complex
inference algorithms.
We consider the \gls{VAE} setup from \Cref{fig:vae} and the binarized
MNIST data set \citep{salakhutdinov2008quantitative}.  We use $d=50$
latent variables per data point and optimize using ADAM.  We study
different components of the \gls{VAE} setup using different methods;
\Cref{appendix:vae} is a complete script.  After training we evaluate
held-out log likelihoods, which are lower bounds on the true value.

\Cref{table:mnist} shows the results.  The first method uses the
\gls{VAE} from \Cref{fig:vae}. The next three methods use the same
\gls{VAE} but apply different gradient estimators: reparameterization
gradient without an analytic KL; reparameterization gradient with an
analytic entropy; and the score function gradient
\citep{paisely2012variational,ranganath:2014}.  This typically leads
to the same optima but at different convergence rates. The score
function gradient was slowest. Gradients with an analytic entropy
produced difficulties around convergence: we switched to stochastic
estimates of the entropy as it approached an optima. We also use
\glspl{HVM} \citep{ranganath2016hierarchical} with a normalizing flow
prior; it produced similar results as a normalizing flow on the latent
variable space \citep{rezende2015variational}, and better than
\glspl{IWAE} \citep{burda2016importance}.

We also study novel combinations, such as \glspl{HVM} with the
\acrshort{IWAE} objective, \gls{GAN}-based optimization on the
decoder (with pixel intensity-valued data),
and R\'{e}nyi divergence on the decoder.  \gls{GAN}-based
optimization does not enable calculation of the log-likelihood;
R\'{e}nyi divergence does not directly optimize for log-likelihood so
it does not perform well. The key point is that
Edward is a convenient research platform:
they are all easy modifications of a given script.

\subsection{GPU-accelerated Hamiltonian Monte Carlo}
\label{sub:gpu}

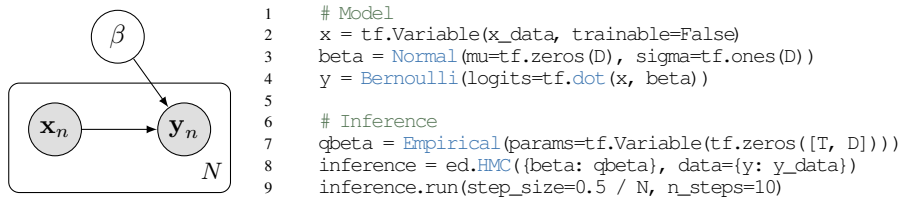
\begin{figure}[!htb]
\vspace{-1.5ex}
\begin{subfigure}{0.3\columnwidth}
  \centering
  \begin{tikzpicture}

  \node[latent]              (beta)      {$\beta$} ;
  \node[right=0.4cm of beta]               (betaph)      {} ;
  \node[obs, below=0.75cm of betaph]      (y)    {$\mby_n$} ;
  \node[obs, left=1.0cm of y] (x)    {$\mbx_n$} ;

  \edge{beta}{y};
  \edge{x}{y};

  \plate[inner sep=0.25cm,
    label={[xshift=-15pt,yshift=15pt]south east:$N$}] {plate1} {
    (y)(x)
  } {};

\end{tikzpicture}
\end{subfigure}%
\begin{subfigure}{0.6\columnwidth}
  \centering
\begin{lstlisting}[language=python]
  # Model
  x = tf.Variable(x_data, trainable=False)
  beta = Normal(mu=tf.zeros(D), sigma=tf.ones(D))
  y = Bernoulli(logits=tf.dot(x, beta))

  # Inference
  qbeta = Empirical(params=tf.Variable(tf.zeros([T, D])))
  inference = ed.HMC({beta: qbeta}, data={y: y_data})
  inference.run(step_size=0.5 / N, n_steps=10)
\end{lstlisting}
\end{subfigure}
\caption{Edward program for Bayesian logistic regression with \gls{HMC}.}
\label{fig:logistic_regression}
\vspace{-1.5ex}
\end{figure}

\begin{table}[tb]
\centering
\begin{tabular}{lr}
\toprule
Probabilistic programming system & Runtime (s)
\\
\midrule
Handwritten NumPy (1 CPU) & 534 \\
Stan (1 CPU) \citep{carpenter2016stan} & 171 \\
PyMC3 (12 CPU) \citep{salvatier2015probabilistic} & 30.0 \\
\textbf{Edward (12 CPU)} & \textbf{8.2} \\
Handwritten TensorFlow (GPU) & 5.0 \\
\textbf{Edward (GPU)} & \textbf{4.9}\\
\bottomrule
\end{tabular}
\caption{\gls{HMC} benchmark for large-scale logistic regression.
Edward (GPU) is significantly faster than other systems. In addition,
Edward has no overhead: it is as fast as handwritten TensorFlow.}
\label{table:hmc}
\end{table}

We benchmark runtimes for
a fixed number of Hamiltonian Monte Carlo \citep[\gls{HMC};][]{neal2011mcmc}
iterations on modern
hardware: a 12-core Intel i7-5930K CPU
at 3.50GHz and an NVIDIA Titan X (Maxwell) GPU. We apply
logistic regression on the
Covertype dataset ($N=581012$, $D=54$; responses were
binarized)
using Edward,
Stan (with PyStan) \citep{carpenter2016stan}, and PyMC3
\citep{salvatier2015probabilistic}.
We ran 100 \gls{HMC} iterations,
with 10 leapfrog updates per iteration, a step size of $0.5 / N$, and
single precision.
\Cref{fig:logistic_regression} illustrates the program in Edward.

\Cref{table:hmc} displays the runtimes.%
\footnote{In a previous version of this paper, we reported PyMC3 took 361s. This was
caused by a bug preventing PyMC3 from correctly handling
single-precision floating point. (PyMC3 with double precision is roughly 14x slower than Edward (GPU).) This has been fixed
after discussion with Thomas Wiecki. The reported numbers also
exclude compilation time, which is significant for Stan.}
Edward (GPU) features a dramatic 35x speedup over Stan
(1 CPU) and 6x speedup over PyMC3 (12 CPU).
This showcases the value of building a \gls{PPL} on top of
computational graphs.
The speedup stems from fast matrix multiplication
when calculating the model's log-likelihood; GPUs can efficiently
parallelize this computation.
We expect similar speedups for models whose bottleneck is also matrix multiplication, such as deep neural networks.

There are various reasons for the speedup. Stan only used 1
CPU as it leverages multiple cores by running \gls{HMC} chains in
parallel. Stan also used double-precision floating point as it
does not allow single-precision. For PyMC3, we note Edward's speedup is
not a result of PyMC3's Theano backend compared to Edward's
TensorFlow. Rather, PyMC3 does not use Theano for all its
computation, so it experiences communication overhead with NumPy.
(PyMC3 was actually slower when using the GPU.)
We predict that porting Edward's design to Theano would feature
similar speedups.

In addition to these speedups, we highlight that Edward has no runtime
overhead: it is as fast as handwritten TensorFlow.  Following
\Cref{sub:inference}, this is because the computational graphs for
inference are in fact the same for Edward and the handwritten code.

\subsection{Probability Zoo}

In addition to Edward, we also release the \emph{Probability Zoo}, a
community repository for pre-trained probability models and their
posteriors.\footnote{%
  The Probability Zoo is available at \url{http://edwardlib.org/zoo}.
  It includes model parameters and inferred posterior factors, such as
  local and global variables during training and any inference
  networks.  } It is inspired by the model zoo in Caffe
\citep{jia2014caffe}, which provides many pre-trained discriminative
neural networks, and which has been key to making large-scale deep learning
more transparent and accessible. It is also inspired by Forest
\citep{stuhlmueller2012forest}, which provides examples of
probabilistic programs.

\section{Discussion: Challenges \& Extensions}
\label{sec:discussion}

We described Edward, a
Turing-complete \gls{PPL}
with compositional representations for probabilistic models and
inference.
Edward expands the scope of probabilistic programming to be as
flexible and
computationally efficient as traditional deep learning.
For flexibility, we showed how Edward can
use a variety of composable inference methods,
capture recent advances in
variational inference and \acrlongpl{GAN}, and
finely control the inference algorithms.
For efficiency, we showed how Edward leverages computational graphs to
achieve fast, parallelizable computation, scales to massive data, and
incurs no runtime overhead over handwritten code.

In present work, we are applying Edward as a research platform
for developing new probabilistic models
\citep{rudolph2016exponential,tran2017deep} and new
inference algorithms \citep{dieng2016chi}.
As with any language design, Edward makes tradeoffs in pursuit
of its flexibility and speed for research. For
example, an open challenge in Edward is to better facilitate programs
with complex control flow and recursion. While possible to represent,
it is unknown how to enable their flexible inference strategies. In
addition, it is open how to expand Edward's design to dynamic
computational graph frameworks---which provide more flexibility in
their programming paradigm---but may sacrifice performance. A crucial
next step for probabilistic programming is to leverage dynamic
computational graphs while maintaining the flexibility and
efficiency that Edward offers.

\subsubsection*{Acknowledgements}
We thank the probabilistic programming community---for sharing our
enthusiasm and motivating further work---including developers of
Church, Venture, Gamalon, Hakaru, and WebPPL.  We also thank Stan developers
for providing extensive feedback as we
developed the language, as well as Thomas Wiecki for experimental
details.
We thank the Google BayesFlow team---Joshua Dillon, Ian Langmore,
Ryan Sepassi, and Srinivas Vasudevan---as well as Amr Ahmed, Matthew
Johnson, Hung Bui, Rajesh Ranganath, Maja Rudolph, and Francisco Ruiz
for their helpful feedback.
This work is supported by NSF IIS-1247664, ONR N00014-11-1-0651,
DARPA FA8750-14-2-0009, DARPA N66001-15-C-4032, Adobe, Google, NSERC
PGS-D, and the Sloan Foundation.

\bibliographystyle{iclr2017_conference}
\bibliography{bib}
\clearpage

\appendix
\section{Model Examples}
\label{appendix:model}

There are many examples available at \url{http://edwardlib.org},
including models, inference methods, and complete scripts.
Below we describe several model examples; \Cref{appendix:svi}
describes an inference example (stochastic variational inference);
\Cref{appendix:complete} describes complete scripts.
All examples in this paper are comprehensive, only leaving out import
statements and fixed values.
See the companion webpage for this paper
(\url{http://edwardlib.org/iclr2017}) for examples in a
machine-readable format with
runnable code.

\subsection{Bayesian Neural Network for Classification}
\label{appendix:bnn}

A Bayesian neural network is a neural network with a prior
distribution on its weights.

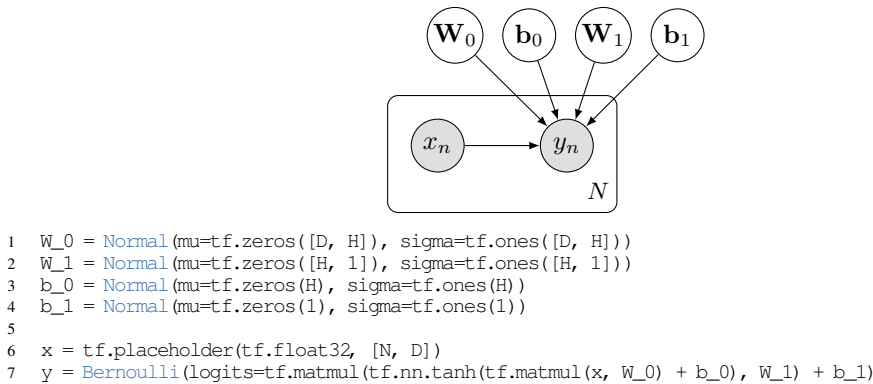
\begin{figure}[tb]
\begin{subfigure}{\columnwidth}
  \centering
  \begin{tikzpicture}

  \node[latent]              (W0)      {$\mbW_0$} ;
  \node[latent, right=of W0, xshift=-0.75cm]              (b0)      {$\mbb_0$} ;
  \node[latent, right=of b0, xshift=-0.75cm]              (W1)      {$\mbW_1$} ;
  \node[latent, right=of W1, xshift=-0.75cm]              (b1)      {$\mbb_1$} ;
  \node[obs, below=0.75cm of b0, xshift=0.5cm]      (y)    {$y_n$} ;
  \node[obs, left=1.0cm of y] (x)    {$x_n$} ;

  \edge{W0}{y};
  \edge{b0}{y};
  \edge{W1}{y};
  \edge{b1}{y};
  \edge{x}{y};

  \plate[inner sep=0.3cm,
    label={[xshift=-15pt,yshift=15pt]south east:$N$}] {plate1} {
    (y)(x)
  } {};

\end{tikzpicture}
\end{subfigure}%
\\
\begin{subfigure}{\columnwidth}
\begin{lstlisting}[language=python]
W_0 = Normal(mu=tf.zeros([D, H]), sigma=tf.ones([D, H]))
W_1 = Normal(mu=tf.zeros([H, 1]), sigma=tf.ones([H, 1]))
b_0 = Normal(mu=tf.zeros(H), sigma=tf.ones(H))
b_1 = Normal(mu=tf.zeros(1), sigma=tf.ones(1))

x = tf.placeholder(tf.float32, [N, D])
y = Bernoulli(logits=tf.matmul(tf.nn.tanh(tf.matmul(x, W_0) + b_0), W_1) + b_1)
\end{lstlisting}
\end{subfigure}
\caption{Bayesian neural network for classification.}
\label{fig:bnn}
\end{figure}

Define the likelihood of an observation $(\mathbf{x}_n, y_n)$ with
binary label $y_n\in\{0,1\}$ as
\begin{align*}
  p(y_n \mid \mbW_0, \mbb_0, \mbW_1, \mbb_1 \;;\; \mathbf{x}_n)
  &=
  \operatorname{Bernoulli}(y_n \g \mathrm{NN}(\mathbf{x}_n\;;\;
  \mbW_0, \mbb_0, \mbW_1, \mbb_1)),
\end{align*}
where $\mathrm{NN}$ is a 2-layer neural network whose weights and biases form
the latent variables $\mbW_0, \mbb_0, \mbW_1, \mbb_1$.
Define the prior on the weights and biases to be the standard normal.
See \Cref{fig:bnn}. There are $N$ data points, $D$ features, and $H$
hidden units.

\subsection{Latent Dirichlet Allocation}
\label{appendix:lda}

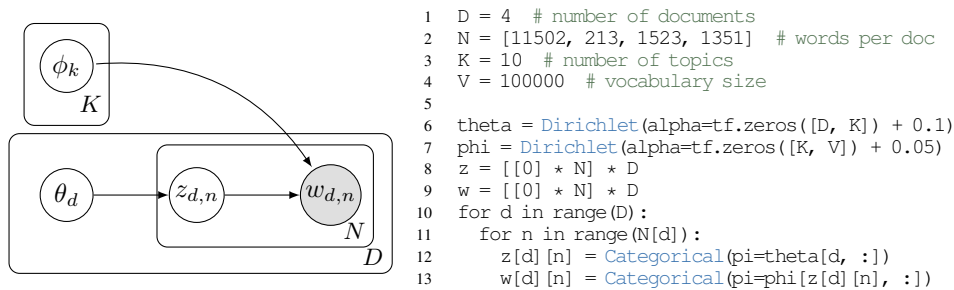
\begin{figure}[!h]
\begin{subfigure}{0.45\columnwidth}
  \centering
  \begin{tikzpicture}

  \node[latent]               (phi)      {$\phi_k$} ;
  \node[latent, below=1.0cm of phi] (theta)    {$\theta_d$} ;
  \node[latent, right=1.0cm of theta] (z)    {$z_{d,n}$} ;
  \node[obs, right=1.0cm of z]      (w)        {$w_{d,n}$} ;

  \draw[style=arrow, bend left] (phi) to (w);
  \edge{z}{w};
  \edge{theta}{z};

  \plate[inner sep=0.15cm, yshift=0.1cm,
    label={[xshift=-15pt,yshift=13pt]south east:$N$}] {plate1} {
    (w)(z)
  } {};
  \plate[inner sep=0.4cm,
    label={[xshift=-15pt,yshift=13pt]south east:$D$}] {plate2} {
    (w)(z)(theta)
  } {};
  \plate[inner sep=0.2cm,
    label={[xshift=-15pt,yshift=15pt]south east:$K$}] {plate3} {
    (phi)
  } {};

\end{tikzpicture}
\end{subfigure}%
\begin{subfigure}{0.55\columnwidth}
\begin{lstlisting}[language=python]
D = 4  # number of documents
N = [11502, 213, 1523, 1351]  # words per doc
K = 10  # number of topics
V = 100000  # vocabulary size

theta = Dirichlet(alpha=tf.zeros([D, K]) + 0.1)
phi = Dirichlet(alpha=tf.zeros([K, V]) + 0.05)
z = [[0] * N] * D
w = [[0] * N] * D
for d in range(D):
  for n in range(N[d]):
    z[d][n] = Categorical(pi=theta[d, :])
    w[d][n] = Categorical(pi=phi[z[d][n], :])
\end{lstlisting}
\end{subfigure}
\caption{Latent Dirichlet allocation \citep{blei2003latent}.}
\label{fig:lda}
\end{figure}

See \Cref{fig:lda}. Note that the program is written for illustration. We
recommend vectorization in practice: instead of storing scalar random
variables in lists of lists, one should prefer to represent few random
variables, each which have many dimensions.

\subsection{Gaussian Matrix Factorizationn}
\label{appendix:gaussian_mf}

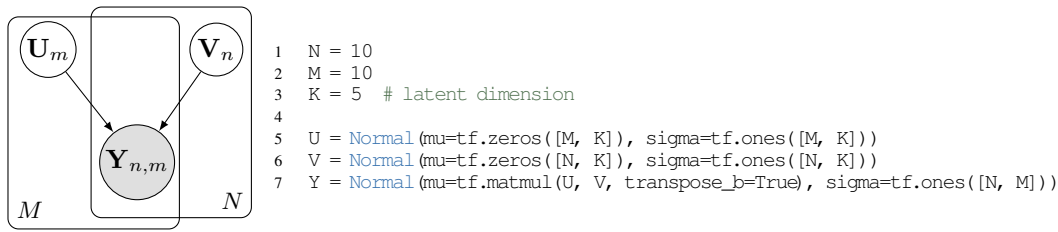
\begin{figure}[!h]
\begin{subfigure}{0.3\columnwidth}
  \centering
  \begin{tikzpicture}

  \node[latent]               (U)      {$\mbU_{m}$} ;
  \node[obs, right=0.3cm of U, yshift=-1.5cm] (Y)    {$\mbY_{n,m}$} ;
  \node[latent, right=1.5cm of U] (V)    {$\mbV_{n}$} ;

  \edge{U}{Y};
  \edge{V}{Y};

  \plate[inner sep=0.1cm,yshift=-0.05cm,xshift=-0.05cm,
    label={[xshift=17pt,yshift=14pt]south west:$M$}] {plate1} {
    (U)(Y)
  } {};
  \plate[inner sep=0.1cm,yshift=0.1cm,
    label={[xshift=-15pt,yshift=13pt]south east:$N$}] {plate2} {
    (V)(Y)
  } {};

\end{tikzpicture}
\end{subfigure}%
\begin{subfigure}{0.7\columnwidth}
\begin{lstlisting}[language=python]
N = 10
M = 10
K = 5  # latent dimension

U = Normal(mu=tf.zeros([M, K]), sigma=tf.ones([M, K]))
V = Normal(mu=tf.zeros([N, K]), sigma=tf.ones([N, K]))
Y = Normal(mu=tf.matmul(U, V, transpose_b=True), sigma=tf.ones([N, M]))
\end{lstlisting}
\end{subfigure}
\caption{Gaussian matrix factorization.}
\label{fig:gaussian_mf}
\end{figure}

See \Cref{fig:gaussian_mf}.

\subsection{Dirichlet Process Mixture Model}
\label{appendix:dirichlet_process}

See \Cref{fig:dp}.

\begin{figure}[!h]
A Dirichlet process mixture model is written as follows:
\begin{lstlisting}[language=python]
mu = DirichletProcess(alpha=0.1, base_cls=Normal, mu=tf.zeros(D), sigma=tf.ones(D), sample_n=N)
x = Normal(mu=mu, sigma=tf.ones([N, D]))
\end{lstlisting}
where \texttt{mu} has shape \texttt{(N, D)}. The
\texttt{DirichletProcess} random variable returns \texttt{sample_n=N}
draws, each with shape given by the base distribution
\texttt{Normal(mu, sigma)}.
The essential component defining the \texttt{DirichletProcess} random
variable is a stochastic while loop. We define it below. See Edward's code
base for a more involved version with a base distribution.
\begin{lstlisting}[language=python]
def dirichlet_process(alpha):
  def cond(k, beta_k):
    flip = Bernoulli(p=beta_k)
    return tf.equal(flip, tf.constant(1))

  def body(k, beta_k):
    beta_k = beta_k * Beta(a=1.0, b=alpha)
    return k + 1, beta_k

  k = tf.constant(0)
  beta_k = Beta(a=1.0, b=alpha)
  stick_num, stick_beta = tf.while_loop(cond, body, loop_vars=[k, beta_k])
  return stick_num
\end{lstlisting}
\caption{Dirichlet process mixture model.}
\label{fig:dp}
\end{figure}

\section{Inference Example: Stochastic Variational Inference}
\label{appendix:svi}

In the subgraph setting, we do data subsampling while working with a
subgraph of the full model. This setting is necessary when the data
and model do not fit in memory.
It is scalable in that both the
algorithm's computational complexity (per iteration) and memory
complexity are independent of the data set size.

For the code, we use the running example, a mixture model
described in \Cref{fig:hierarchical_model_example}.
\begin{lstlisting}[language=Python]
N = 10000000  # data set size
D = 2  # data dimension
K = 5  # number of clusters
\end{lstlisting}
The model is
\begin{equation*}
p(\mbx, \mathbf{z}, \beta)
= p(\beta) \prod_{n=1}^N p(z_n \mid \beta) p(x_n \mid z_n, \beta).
\end{equation*}
To avoid memory issues, we work on only a subgraph of the model,
\begin{equation*}
p(\mbx, \mathbf{z}, \beta)
= p(\beta) \prod_{m=1}^M p(z_m \mid \beta) p(x_m \mid z_m, \beta)
\end{equation*}
\begin{lstlisting}[language=Python]
M = 128  # mini-batch size

beta = Normal(mu=tf.zeros([K, D]), sigma=tf.ones([K, D]))
z = Categorical(logits=tf.zeros([M, K]))
x = Normal(mu=tf.gather(beta, z), sigma=tf.ones([M, D]))
\end{lstlisting}
Assume the variational model is
\begin{equation*}
q(\mathbf{z}, \beta) =
q(\beta; \lambda) \prod_{n=1}^N q(z_n \mid \beta; \gamma_n),
\end{equation*}
parameterized by $\{\lambda, \{\gamma_n\}\}$.
Again, we work on only a subgraph of the model,
\begin{equation*}
q(\mathbf{z}, \beta) =
q(\beta; \lambda) \prod_{m=1}^M q(z_m \mid \beta; \gamma_m).
\end{equation*}
parameterized by $\{\lambda, \{\gamma_m\}\}$. Importantly, only $M$
parameters are stored in memory for $\{\gamma_m\}$ rather than $N$.
\begin{lstlisting}[language=Python]
qbeta = Normal(mu=tf.Variable(tf.zeros([K, D])),
               sigma=tf.nn.softplus(tf.Variable(tf.zeros[K, D])))
qz_variables = tf.Variable(tf.zeros([M, K]))
qz = Categorical(logits=qz_variables)
\end{lstlisting}
We use \texttt{KLqp}, a variational method that minimizes
the divergence measure $\operatorname{KL}(q\gg p)$
\citep{jordan1999introduction}.
We instantiate two algorithms: a global inference over $\beta$ given
the subset of $\mathbf{z}$ and a local inference over the subset of
$\mathbf{z}$ given $\beta$.
We also pass in a TensorFlow placeholder \texttt{x_ph} for the data,
so we can change the data at each step.
\begin{lstlisting}[language=Python]
x_ph = tf.placeholder(tf.float32, [M])
inference_global = ed.KLqp({beta: qbeta}, data={x: x_ph, z: qz})
inference_local = ed.KLqp({z: qz}, data={x: x_ph, beta: qbeta})
\end{lstlisting}
We initialize the algorithms with the \texttt{scale} argument, so that
computation on \texttt{z} and \texttt{x} will be scaled appropriately.
This enables unbiased estimates for stochastic gradients.
\begin{lstlisting}[language=Python]
inference_global.initialize(scale={x: float(N) / M, z: float(N) / M})
inference_local.initialize(scale={x: float(N) / M, z: float(N) / M})
\end{lstlisting}
We now run the algorithm, assuming there is a \texttt{next_batch}
function which provides the next batch of data.
\begin{lstlisting}[language=Python]
qz_init = tf.initialize_variables([qz_variables])
for _ in range(1000):
  x_batch = next_batch(size=M)
  for _ in range(10):  # make local inferences
    inference_local.update(feed_dict={x_ph: x_batch})

  # update global parameters
  inference_global.update(feed_dict={x_ph: x_batch})
  # reinitialize the local factors
  qz_init.run()
\end{lstlisting}
After each iteration, we also reinitialize the parameters for
$q(\mathbf{z}\mid\beta)$; this is because we do inference on a new
set of local variational factors for each batch.
This demo readily applies to other inference algorithms such as
\texttt{SGLD} (stochastic gradient Langevin dynamics): simply
replace \texttt{qbeta} and \texttt{qz} with \texttt{Empirical} random
variables; then call \texttt{ed.SGLD} instead of \texttt{ed.KLqp}.

Note that if the data and model fit in memory but you'd still like to
perform data subsampling for fast inference, we recommend not defining
subgraphs. You can reify the full model, and simply index the local
variables with a placeholder. The placeholder is fed at runtime to
determine which of the local variables to update at a time. (For more
details, see the website's API.)

\section{Complete Examples}
\label{appendix:complete}

\subsection{Variational Auto-encoder}
\label{appendix:vae}

See \Cref{fig:appendix_vae}.

\begin{figure}[!h]
\begin{lstlisting}[language=python]
import edward as ed
import tensorflow as tf

from edward.models import Bernoulli, Normal
from scipy.misc import imsave
from tensorflow.contrib import slim
from tensorflow.examples.tutorials.mnist import input_data

M = 100  # batch size during training
d = 2  # latent variable dimension

# Probability model (subgraph)
z = Normal(mu=tf.zeros([M, d]), sigma=tf.ones([M, d]))
h = Dense(256, activation='relu')(z)
x = Bernoulli(logits=Dense(28 * 28, activation=None)(h))

# Variational model (subgraph)
x_ph = tf.placeholder(tf.float32, [M, 28 * 28])
qh = Dense(256, activation='relu')(x_ph)
qz = Normal(mu=Dense(d, activation=None)(qh),
            sigma=Dense(d, activation='softplus')(qh))

# Bind p(x, z) and q(z | x) to the same TensorFlow placeholder for x.
mnist = input_data.read_data_sets("data/mnist", one_hot=True)
data = {x: x_ph}

inference = ed.KLqp({z: qz}, data)
optimizer = tf.train.RMSPropOptimizer(0.01, epsilon=1.0)
inference.initialize(optimizer=optimizer)

tf.initialize_all_variables().run()

n_epoch = 100
n_iter_per_epoch = 1000
for _ in range(n_epoch):
  for _ in range(n_iter_per_epoch):
    x_train, _ = mnist.train.next_batch(M)
    info_dict = inference.update(feed_dict={x_ph: x_train})

  # Generate images.
  imgs = x.value().eval()
  for m in range(M):
    imsave("img/%d.png" % m, imgs[m].reshape(28, 28))
\end{lstlisting}
\caption{Complete script for a \gls{VAE}
\citep{kingma2014autoencoding} with batch training.
It generates MNIST digits after every 1000 updates.}
\label{fig:appendix_vae}
\end{figure}

\subsection{Probabilistic Model for Word Embeddings}
\label{appendix:ef_emb}

\begin{figure}[!h]
\begin{lstlisting}[language=python]
import edward as ed
import tensorflow as tf

from edward.models import Bernoulli, Normal, PointMass

N = 581238  # number of total words
M = 128  # batch size during training
K = 100  # number of factors
ns = 3  # number of negative samples
cs = 4  # context size
L = 50000  # vocabulary size

# Prior over embedding vectors
p_rho = Normal(mu=tf.zeros([M, K]),
               sigma=tf.sqrt(N) * tf.ones([M, K]))
n_rho = Normal(mu=tf.zeros([M, ns, K]),
               sigma=tf.sqrt(N) * tf.ones([M, ns, K]))

# Prior over context vectors
ctx_alphas = Normal(mu=tf.zeros([M, cs, K]),
                    sigma=tf.sqrt(N)*tf.ones([M, cs, K]))

# Conditional likelihoods
ctx_sum = tf.reduce_sum(ctx_alphas, [1])
p_eta = tf.expand_dims(tf.reduce_sum(p_rho * ctx_sum, -1),1)
n_eta = tf.reduce_sum(n_rho * tf.tile(tf.expand_dims(ctx_sum, 1), [1, ns, 1]), -1)
y_pos = Bernoulli(logits=p_eta)
y_neg = Bernoulli(logits=n_eta)

# placeholders for batch training
p_idx = tf.placeholder(tf.int32, [M, 1])
n_idx = tf.placeholder(tf.int32, [M, ns])
ctx_idx = tf.placeholder(tf.int32, [M, cs])

# Variational parameters (embedding vectors)
rho_params = tf.Variable(tf.random_normal([L, K]))
alpha_params = tf.Variable(tf.random_normal([L, K]))

# Variational distribution on embedding vectors
q_p_rho = PointMass(params=tf.squeeze(tf.gather(rho_params, p_idx)))
q_n_rho = PointMass(params=tf.gather(rho_params, n_idx))
q_alpha = PointMass(params=tf.gather(alpha_params, ctx_idx))

inference = ed.MAP(
  {p_rho: q_p_rho, n_rho: q_n_rho, ctx_alphas: q_alpha},
  data={y_pos: tf.ones((M, 1)), y_neg: tf.zeros((M, ns))})

inference.initialize()
tf.initialize_all_variables().run()

for _ in range(inference.n_iter):
  targets, windows, negatives = next_batch(M)  # a function to generate data
  info_dict = inference.update(feed_dict={p_idx: targets, ctx_idx: windows, n_idx: negatives})
  inference.print_progress(info_dict)
\end{lstlisting}
\caption{Exponential family embedding for binary data \citep{rudolph2016exponential}. Here, \gls{MAP}
is used to maximize the total sum of conditional log-likelihoods and log-priors.}
\label{fig:ef_emb}
\end{figure}

See \Cref{fig:ef_emb}.
This example uses data subsampling (\Cref{sub:batch_training}). The
priors and conditional likelihoods are defined only for a minibatch of
data. Similarly the variational model only models the embeddings used
in a given minibatch. TensorFlow variables contain the embedding
vectors for the entire vocabulary. TensorFlow placeholders ensure that
the correct embedding vectors are used as variational parameters for a
given minibatch.

The Bernoulli variables \texttt{y_pos} and \texttt{y_neg} are fixed to
be $1$'s and $0$'s respectively. They model whether a word is indeed the target word
for a given context window or has been drawn as a negative sample.
Without regularization (via priors), the objective we optimize is
identical to negative sampling.

\end{document}